%% file: main.tex
\documentclass{article}

\usepackage[preprint, nonatbib]{neurips_2022}

\usepackage[utf8]{inputenc} 
\usepackage[T1]{fontenc}    
\usepackage{hyperref}       
\usepackage{url}            
\usepackage{booktabs}       
\usepackage{amsfonts}       
\usepackage{nicefrac}       
\usepackage{microtype}      
\usepackage{xcolor}         
\usepackage[pdftex]{graphicx}
\usepackage{mathtools}
\usepackage{textgreek}

\newcommand{\partitle}[1]{\bigbreak\noindent\textbf{#1}}

\title{Evaluating Continual Test-Time Adaptation for Contextual and Semantic Domain Shifts}

\author{%
  Tommie Kerssies\\
  Eindhoven University of Technology\\
  \texttt{t.kerssies@student.tue.nl}\\
    \And
  Mert Kılıçkaya\\
  Eindhoven University of Technology\\
  \texttt{m.kilickaya@tue.nl}\\
  \And
  Joaquin Vanschoren\\
  Eindhoven University of Technology\\
  \texttt{j.vanschoren@tue.nl}\\
}

\begin{document}

\maketitle

\begin{abstract}


In this paper, our goal is to adapt a pre-trained convolutional neural network to domain shifts at test time. We do so continually with the incoming stream of test batches, without labels. The existing literature mostly operates on artificial shifts obtained via adversarial perturbations of a test image. 
Motivated by this, we evaluate the state of the art on two realistic and challenging sources of domain shifts, namely contextual and semantic shifts. 
Contextual shifts correspond to the environment types, for example, a model pre-trained on indoor context has to adapt to the outdoor context on CORe-$50$~\cite{lomonaco2017core50}. Semantic shifts correspond to the capture types, for example a model pre-trained on natural images has to adapt to cliparts, sketches, and paintings on DomainNet~\cite{peng2019moment}. We include in our analysis recent techniques such as Prediction-Time Batch Normalization (BN)~\cite{nado2020evaluating}, Test Entropy Minimization (TENT)~\cite{wang2020tent} and Continual Test-Time Adaptation (CoTTA)~\cite{wang2022continual}. Our findings are three-fold: \textit{i)} Test-time adaptation methods perform better and forget less on contextual shifts compared to semantic shifts, \textit{ii)} TENT outperforms other methods on short-term adaptation, whereas CoTTA outpeforms other methods on long-term adaptation, \textit{iii)} BN is most reliable and robust. Our code is available at \href{https://github.com/tommiekerssies/Evaluating-Continual-Test-Time-Adaptation-for-Contextual-and-Semantic-Domain-Shifts}{https://github.com/tommiekerssies/Evaluating-Continual-Test-Time-Adaptation-for-Contextual-and-Semantic-Domain-Shifts}.

\end{abstract}

\input{0-intro}
\input{1-relwork}
\input{2-method}
\input{3-setup}
\input{4-analysis}
\input{5-discussion}
\input{6-conclusion}
\input{7-futurework}

\bibliographystyle{abbrv}  
\bibliography{references}




\end{document}

%% file: 0-intro.tex
\section{Introduction}
\partitle{Motivation for domain adaptation.}
Distributional shifts in evaluation data alter a model's performance.
Domain adaptation is proposed to adapt an existing model to novel situations. More specifically, domain adaptation adapts a model trained on one or more source domains to a target domain. Regular domain adaptation considers a stationary target, meaning that the target does not change over time. 

\partitle{Motivation for continual learning.}
Continual learning is a concept to learn a model on data with continually shifting distributions or even changing tasks, without forgetting what was learned previously. Storing data may not be feasible due to storage or privacy concerns, therefore we only assume access to new data in the sequence. In a domain adaptation context, having access to the entire target dataset a priori may not be realistic, because real-world data is collected continually and the distribution may shift over time.

\partitle{Continual domain adaptation.}
To address the need for a model that can adapt over time, continual domain adaptation is proposed. More specifically, this work focuses on continual test-time adaptation, where we do not assume access to the source data and adapt online to continually arriving unlabeled test batches.

\partitle{Contextual and semantic domain shifts.}
This work focuses on visual recognition tasks (image classification). We distinguish two types of distribution shifts: contextual and semantic shifts. In a semantic distribution shift, the semantics of the relevant parts of the image change (e.g. an object is captured in real life instead of a drawing). In a contextual shift, the semantics of the relevant parts of the image remain the same, but may appear differently due to a context change (e.g. the lighting conditions change).

\partitle{Evaluation of continual domain adaptation in contextual and semantic domain shifts.}
We evaluate Prediction-Time Batch Normalization (BN)~\cite{nado2020evaluating}, Test Entropy Minimization (TENT)~\cite{wang2020tent} and Continual Test-Time Adaptation (CoTTA)~\cite{wang2022continual} for continual test-time adaptation to contextual distribution shifts on CORe50~\cite{lomonaco2017core50} and semantic distribution shifts on DomainNet~\cite{peng2019moment}. We find that the test-time adaptation methods show relatively more improvements and less forgetting on contextual shifts compared to semantic shifts. For short-term adaptation, we find that TENT performs best and for long-term adaptation, we find that CoTTA performs best. Finally, we find that unlike TENT or CoTTA, BN does not show error accumulation or catastrophic forgetting, and works best when the source model was trained on a dataset with a relatively narrow distribution. 

%% file: 1-relwork.tex
\section{Related Work}
\subsection{Transfer Learning}
Transfer learning is about transferring knowledge from a source to a target. Transfer learning is a very broad term and can be divided into different subsettings. Following the work of \cite{redko2020survey}, we consider different scenarios. First of all, whether the source and target data are from the same distribution or not. If source and target are from the same distribution, and the task of source and target is the same, this is the usual learning setting. If source and target are from the same distribution, but their task differs, we call this inductive transfer learning. Similarly, the distributions may differ but the task may be the same, which is called transductive transfer learning. Finally, both the distribution and task may be different, called unsupervised transfer learning.\\

\subsection{Domain Adaptation}
Transductive transfer learning, where source and target distributions differ but their task is the same, is more commonly referred to as domain adaptation. Table \ref{da-settings} outlines related settings of domain adaptation with their similarities and differences. 

\begin{table}[h]
  \caption{Settings of domain adaptation.}
  \label{da-settings}
  \centering
  \begin{tabular}{lllll}
    \toprule
    Setting & Source access & Target labels & Target stationary \\
    \midrule
    Supervised domain adaptation & \checkmark & \checkmark & \checkmark \\
    Unsupervised domain adaptation & \checkmark &  & \checkmark \\
    Test-time adaptation & & & \checkmark \\
    Continual test-time adaptation & & & \\
    \bottomrule
  \end{tabular}
\end{table}

\subsubsection{Test-Time Adaptation} 
Test-time adaptation is similar to unsupervised domain adaptation, where no target labels are available. However, unsupervised domain adaptation requires access to the source data when adapting to the target data. Test-time adaptation methods do not require this access, and can therefore adapt to target data at test-time. Table \ref{tta-methods} outlines methods for test-time adaptation that we will evaluate in this work in the setting where the target is continually changing and not stationary.

Test-time adaptation methods can be divided into calibration or optimization based methods. Calibration based methods only update the batch normalization statistics \cite{ioffe2015batch}, whereas optimization based methods will update the learnable parameters in the model with some unsupervised objective.

\partitle{Calibration based.} The concurrent works of \cite{schneider2020improving} and \cite{nado2020evaluating} investigate the effect of updating the statistics at test-time to reduce covariate shift, where \cite{nado2020evaluating} simply proposes to discard the source statistics at test-time and only use test batch statistics. A notable difference with \cite{schneider2020improving} is that \cite{schneider2020improving} proposes to calibrate the batch normalization statistics by combining the source statistics with the test batch statistics using a certain formula with a hyperparameter that can emphasize the source or the target more. Other works, such as \textalpha-BN \cite{you2021test} propose a similar thing. Most of these methods focus mostly on using test batches, however the work of SITA \cite{khurana2021sita} looks into approximating test batch statistics one test image at a time. In the work of \cite{zou2022learning}, the authors propose to learn how to calibrate the batch normalization statistics using a meta-model, which should also work on one test image at a time.

\partitle{Optimization based.}
As test batches do not have labels, an optimization based approach to test-time adaptation needs another objective to optimize for. 

The authors of TENT \cite{wang2020tent} propose to update the batch normalization transformation parameters by minimizing test entropy. CoTTA \cite{wang2022continual} builds on TENT for a continual learning setting, by updating all parameters and introducing weight and augmentation averaged pseudo-labels as well as stochastic restoration. Many other works are also based on TENT and another noteworthy one is \cite{niu2022efficient}, where the authors propose to only minimize entropy when entropy is below a certain threshold, which should minimize error accumulation by minimizing entropy on uncertain samples.

Another approach to optimization based test-time adaptation is self-supervised learning. Test-time training as proposed in \cite{sun2020test} utilizes multi-task learning for a supervised and a self-supervised task when training the source model, and will keep on training the self-supervised task at test-time. A distinction with the setting studied in our work is that this approach will not work on any pre-trained source model and requires the source model to be trained in a certain way. AdaContrast \cite{chen2022contrastive} uses a self-supervised objective based on contrastive learning for test-time adaptation, but should also work in our setting as it does not require a specifically trained source model.

\begin{table}[h]
  \caption{Methods for test-time adaptation.}
  \label{tta-methods}
  \centering
      \begin{tabular}{llll}
        \toprule
        Method & Strategy & Updated parameters & Supervision \\
        \midrule
        BN \cite{nado2020evaluating} & Calibration & $E[x]$, $Var[x]$ & Unsupervised \\
        TENT \cite{wang2020tent} & Calibration, Optimization & $E[x]$, $Var[x]$, $\gamma$, $\beta$ & Pseudo-supervised \\
        CoTTA \cite{wang2022continual} & Calibration, Optimization & All & Pseudo-supervised \\
        \bottomrule
      \end{tabular}
\end{table}

\subsection{Domain Generalization}
In domain generalization terminology, the data a model was trained on is referred to as in-distribution. Data with a distribution different from the training data would be called out-of-distribution. Domain generalization methods aim to train a model to be more generalizable, that is, better performing at out-of-distribution (test) data. Different from domain adaptation, where access to the target domain in some way is required, domain generalization techniques can be applied without having access to any out-of-distribution data. Domain generalization methods can be combined with domain adaptation methods if there is also access to out-of-distribution data. In our work, we do not utilize any domain generalization methods.

%% file: 2-method.tex
\section{Empirical Method}

\partitle{Prediction-time batch normalization (BN).} 
Batch normalization layers in a neural network are shown to make training faster and more stable by normalizing neuron inputs \cite{ioffe2015batch}. The authors of \cite{ioffe2015batch} hypothesize that Batch Normalization layers mitigate the problem of internal covariate shift, which causes performance degradation as well as worsened uncertainty estimation. 

Mathematically, a batch normalization layer can be expressed as:
\begin{equation}
y=\frac{x-\mathrm{E}[x]}{\sqrt{\operatorname{Var}[x]+\epsilon}} * \gamma+\beta
\end{equation}
where $\gamma$ and $\beta$ are learnable transformation parameters and $\mathrm{E}[x]$ and $\operatorname{Var}[x]$ are an estimation of the mean input to the neuron and its variance. These neuron input statistics are commonly estimated during training by using an exponential moving average that is updated for every batch encountered during training:
\begin{equation}
\hat{x}_{\text {t+1}}=\alpha * \hat{x_{t}} + (1-\alpha) * x_{t+1}
\end{equation}
where $\hat{x_{t}}$ is the exponential moving average of the statistic on all previous batches and $x_{t+1}$ is the value of the statistic on the newly encountered batch (and $\alpha$ is a smoothing factor). After training, the estimated neuron input statistics are frozen. Consequently, the frozen neuron input statistics are used for inference.

In case of a distribution shift at test-time, neuron input statistics for the test distribution may be different from the neuron input statistics for the train distribution, causing internal covariate shift, thus performance degradation as well as worsened uncertainty estimation.

In \cite{nado2020evaluating}, the authors propose the test-time adaptation method Prediction-Time Batch Normalization (BN) to do inference on batches of test data and compute the neuron input statistics from the test batch. Instead of using the frozen estimates of neuron input statistics for the training data, the neuron input statistics for the test batch are used. A bigger test batch size will naturally result in better estimates of the neuron input statistics for the true test distribution. BN as defined by \cite{nado2020evaluating} does not have any hyperparameters.

\partitle{Test entropy minimization (TENT).} A visual recognition model for $n$ classes running inference on test image $X$, outputs a vector $Y$ representing probabilities $y_i$ for each possible class $i$. The entropy of this prediction:
\begin{equation}
\mathrm{H}(Y)=-\sum_{i=1}^{n} \mathrm{P}\left(y_{i}\right) \log \mathrm{P}\left(y_{i}\right)
\end{equation}
is a measure for the confidence of the model on its prediction $Y$ for test image $X$. A lower entropy corresponds to a more confident prediction. In \cite{wang2020tent}, the authors find that test entropy correlates with distribution shift, where more entropy correlates with a bigger shift. Based on this finding, they propose the test-time adaptation method Test Entropy Minimization (TENT).

TENT builds on top of BN; inference is done on batches of test data and the neuron input statistics for the test batch are used instead of those estimated during training. The objective in TENT is to minimize test entropy on a batch of test images, by updating the transformation parameters in the batch normalization layers of the model. More specifically, the parameters $\gamma$ and $\beta$ for all batch normalization layers are updated using gradient descent, on every newly encountered test batch. Unlike the neuron input statistics, which are computed while forwarding the test batch through the network, the transformation parameters are updated after the forward pass. The only hyperparameter of TENT is its learning rate.

TENT may suffer from error accumulation. When the model makes an incorrect prediction, TENT will learn from the erroneous prediction. In the next batches, this may lead to the model making more of such errors. As TENT is unsupervised, it cannot recover from this error reinforcement.

\partitle{Continual test-time adaptation (CoTTA).}
In a continually changing test distribution setting, error accumulation may become a significant problem. The authors of \cite{wang2022continual} propose Continual Test-Time Adaptation (CoTTA) to tackle the issues of previous works on test-time adaptation on a continually changing target. CoTTA builds on top of TENT. Instead of only updating the transformation parameters, CoTTA updates all parameters in the model and does not only rely on batch normalization layers for its adaptation. The effects of error accumulation are reduced by updating the model parameters with an exponential moving average:
\begin{equation}
\theta'_{t+1} = \alpha * \theta'_{t} + (1-\alpha) * \theta_{t+1}
\end{equation} 
where a higher value for the smoothing factor $\alpha$ discounts older batches faster, resulting in faster adaptation, but more risk of error accumulation.

Furthermore, with a continually changing target, TENT suffers from catastrophic forgetting. After adapting to the continually changing target, the model may not perform as well anymore on the source distribution. To circumvent this problem, CoTTA stochastically restores source model weights. The restoration hyperparameter $p$ determines the probability that a weight is restored to its original value from the source model. A higher value of $p$ decreases forgetting, but limits the model from adapting.  

%% file: 3-setup.tex
\section{Experimental Setup}
\label{setup}
We evaluate test-time adaptation methods in the continual test-time adaptation setting under contextual and semantic shifts, for which we use CORe50~\cite{lomonaco2017core50} and DomainNet~\cite{peng2019moment} respectively. For both datasets, all images have a label for their class as well as for their domain. Different splits of source and target domains result in different experiments. For each experiment, the source domains are further split into a 90\% train and 10\% validation (hold-out) set. A source model is trained on the train set of the source domains. Whenever we do evaluation on source domain(s) (e.g. to measure forgetting), we do this on the 10\% validation sets only. Whenever we do evaluation on target domain(s), we do this on all the data available for those domains.

For test-time adaptation method CoTTA, we always run it with the default learning rate and disable test-time augmentations.

\subsection{Contextual Shifts on CORe50}
\partitle{Dataset.} For evaluation of test-time adaptation methods under continual contextual shifts, we perform experiments on the CORe50 dataset \cite{lomonaco2017core50}. The dataset consists of 50 domestic objects as classes. These objects are captured in 11 distinct contextually different domains with different backgrounds and lighting conditions. For each object and domain, there are 300 images (frames) from a 15 seconds video of the object. The images are from the point-of-view of the operator moving and rotating the objects in his hand slowly. The holding hand is consistent within a domain and is different between domains. The hand causes the object to be occluded in different ways. The authors of CORe50 did not design the dataset specifically for continual test-time adaptation. Instead, it was designed for supervised continual learning. They propose a split of domains (called sessions in \cite{lomonaco2017core50}) for training and testing. Three of the eleven domains (3, 7 and 8) have been selected for testing and the remaining eight domains have been selected for training. The authors tried to balance the difficulty of the train and test domains with respect to the holding hand, background and indoor/outdoor.

\partitle{Source models.}
Similarly to the source models for DomainNet-126, we train the source models with thee different seeds for each source domain (combination) and evaluate on all three models for each experiment and report the average. The model architecture consists of a feature encoder followed by a fully connected layer as classifier. The feature encoder is based on the ResNet-50~\cite{he2016deep} backbone. The models were initialized with ImageNet-1K~\cite{deng2009imagenet} weights and trained with batch size 128.

\subsection{Semantic Shifts on DomainNet}
\partitle{Dataset.} For evaluation of test-time adaptation methods under continual semantic shifts, we perform experiments on the DomainNet dataset \cite{peng2019moment}. The original dataset consists of 345 classes in six domains. The authors of~\cite{saito2019semi} find that labels of certain classes and domains are very noisy. They propose a less noisy subset of 126 classes and four domains, which we will refer to as DomainNet-126 as they do in \cite{chen2022contrastive}. The domains in DomainNet-126 are real images (18,523 images), paintings (30,032 images), cliparts (18,523 images) and sketches (24,147 images). Although all classes are present in each of the four domains, they have a semantically different appearance.

\partitle{Source models.}
We use the pre-trained models from \cite{chen2022contrastive}. The authors provide models trained with three different seeds for each source domain. We evaluate on all three models for each experiment and report the average. The model architecture consists of a feature encoder followed by a fully connected layer as classifier. The feature encoder is based on the ResNet-50~\cite{he2016deep} backbone. The authors of \cite{chen2022contrastive} follow SHOT \cite{liang2020we} by adding a 256-dimensional bottleneck after the backbone and apply WeightNorm~\cite{salimans2016weight} on the classifier. The models were initialized with ImageNet-1K~\cite{deng2009imagenet} weights and trained with batch size 128.

%% file: 4-analysis.tex
\section{Analysis}
In this section, we will report on the results of the following experiments:

\begin{itemize}
\item [\ref{qualitative-inspection}] A qualitative inspection of the test-time adaptation methods.
\item [\ref{effect-batch-size-on-bn}] An investigation on the effect of batch size on test-time batch normalization statistics adaptation.
\item [\ref{short-term-adaptation}] An evaluation of short-term adaptation performance of the test-time adaptation methods.
\item [\ref{long-term-adaptation}] An evaluation of long-term adaptation performance of the test-time adaptation methods.
\end{itemize}

\subsection{Qualitative Inspection}\label{qualitative-inspection}
\begin{figure}[h]
  \centering
  \includegraphics[width=\columnwidth]{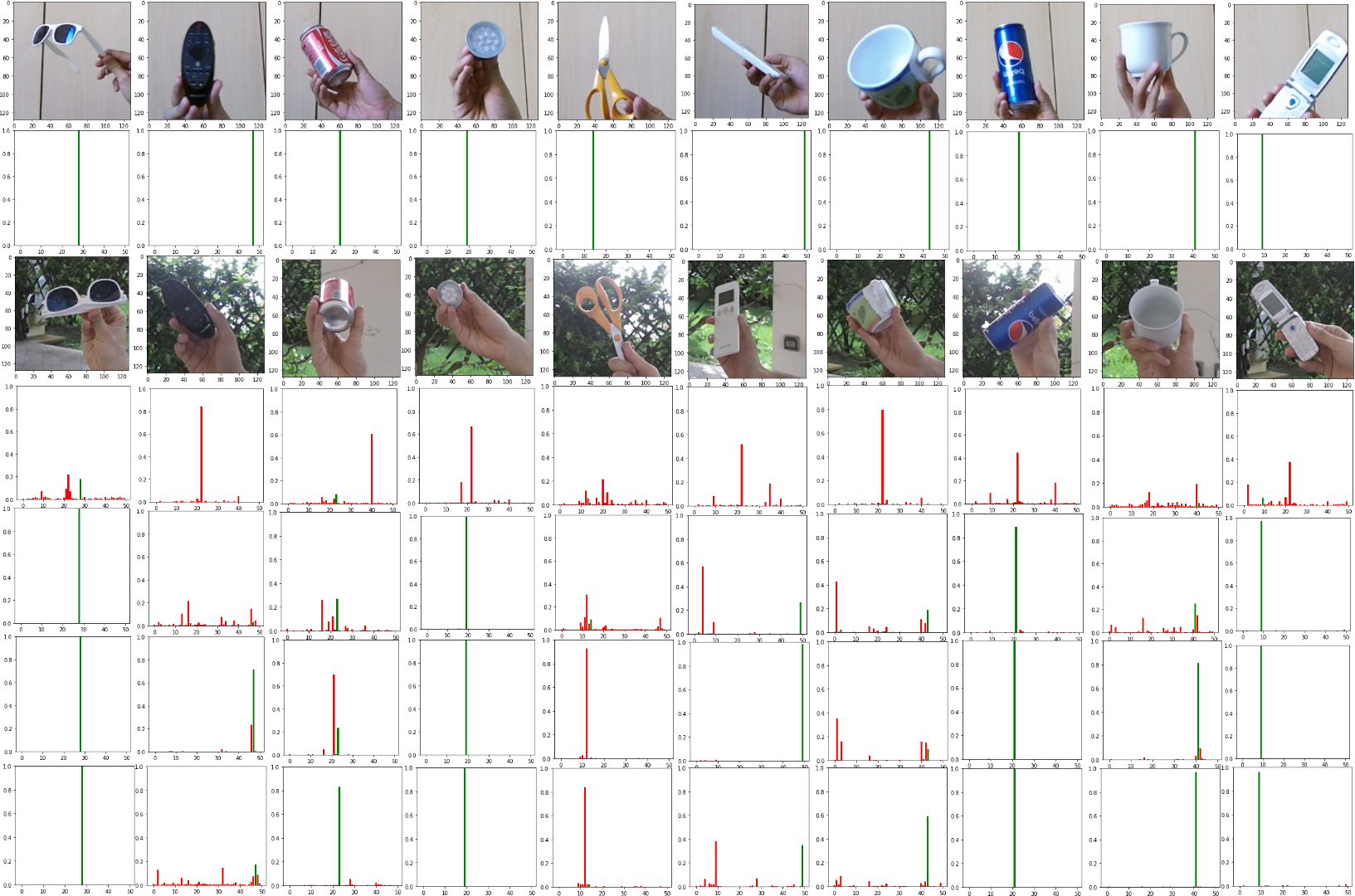}
  \caption{A qualitative inspection of the test-time adaptation methods on 10 example classes from CORe50. The first row shows images of 10 objects from indoor source domain 1. The second row shows the prediction confidence vectors of the source model (10/10 correct). The third row shows the same objects in outdoor target domain 10. The fourth row shows the prediction confidence vectors of the source model on the target domain (0/10 correct). The fifth row shows the prediction confidence vectors of BN on the target domain (6/10 correct). The sixth row shows the prediction confidence vectors of TENT on the target domain (7/10 correct). The last row shows the prediction confidence vectors of CoTTA on the target domain (8/10 correct).}
    \label{fig:qualitative}
\end{figure}

In Figure~\ref{fig:qualitative} we visualize what effect the different methods for test-time adaptation have on the predictions of a model. We do this for 10 example objects from CORe50.

In the top row, an example image is shown for each of the objects in the validation set of domain 1. Below that, in the second row, we visualize the prediction confidence vectors of a model trained only on domain 1 doing inference on these example images. The visualization is a bar chart, where the vertical axis represents prediction confidence and the horizontal axis represents all 50 possible classes. The bar corresponding to the correct class is colored green and all other bars are colored red. The final prediction of the model is simply the tallest bar, so the class with maximal confidence. The source model predicts all objects from the source domain validation set correctly with close to maximum confidence.

In the third row, an example image is shown for each of the objects in target domain 10. This domain is quite different from domain 1, as it is outdoor instead of indoor, resulting in contextual differences in background and lighting. Also the hand might hold the object differently, resulting in different occlusions of the objects. Below, in the fourth row, we see that the source model performs quite poorly due to this domain shift. It doesn't make one correct prediction on domain 10. On top of that, it is quite confident of its mistakes. So the uncertainty estimation of the model is not accurate either.

In the fifth row, we see the results of using BN. Simply updating the batch normalization statistics using a batch of test-time images leads to much more accurate predictions, of which 6 are correct now. The model also has better uncertainty estimation, the incorrect classes (red bars) have lower confidence (smaller bars). Please note that the effective batch size used is 400, so these ten images are part of a bigger batch of images from domain 10.

In the sixth row, we see the results of using TENT. Minimizing test entropy causes the model to make seven correct predictions, one more than BN. Please note that we divided domain 10 into batches of 400 images and that these ten example images are part of the last batch in the sequence. That means that TENT has been updating the batch normalization transformation parameters for many test batches in domain 10 before the predictions we see on the last batch. In general, we see that less classes get a significant confidence (less bars). The classes that do get significant confidence, are also higher in confidence. We see that some incorrect classes are predicted with high confidence, so TENT is making the uncertainty estimation less reliable. By minimizing entropy, TENT caused correct predictions to be more confident, but also accumulated errors by causing incorrect predictions to be more confident. We see TENT 'uncovering' the correct class in image two and six, but we also see TENT 'forgetting' the correct class in image three.

Finally, in the last row, we see the results of CoTTA, which updates all parameters of the network to minimize entropy, but uses weight averaging and stochastic restoration of the source model to reduce error accumulation and catastrophic forgetting. CoTTA causes the model to make eight correct predictions, one more than TENT. Compared to TENT, we see that CoTTA 'uncovered' the correct class in image three and seven, but failed or barely succeeded to 'uncover' the correct class in image two and six (which TENT did better). However, in these failure cases, we do see that the correct class is almost the same confidence as the predicted incorrect class and we hypothesize that after seeing more images from domain 10, these classes may get higher confidence over time.

\subsection{Effect of Batch Size on Test-Time Batch Normalization Statistics Adaptation}
\label{effect-batch-size-on-bn}
\begin{figure}[h]
  \label{bn-batch-size-results}
  \centering
  \includegraphics[width=.7\columnwidth]{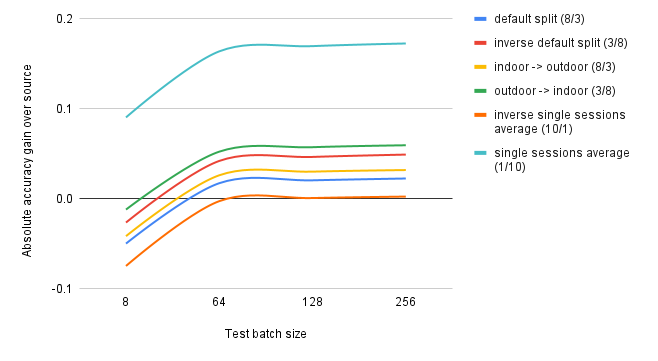}
  \caption{Effect of test batch size on classification accuracy~(\%) using the BN method for models trained on different source domains from CORe50.}
\end{figure}
All methods considered in our work rely on doing inference on batches of test images. Therefore, we investigate the effect of the test batch size on model accuracy when updating batch normalization statistics at test-time on CORe50 using the BN method. We evaluate models trained on different (combinations of) source domains on all 11 domains with and without BN for different batch sizes and we plot the average accuracy gain from BN over source in Figure~\ref{bn-batch-size-results}. These results indicate that with more domains excluded from the source and included in the target, there is more accuracy gain. Furthermore, accuracy gain increases monotonically with batch size, with diminishing returns after some point. In case of a small test batch size, accuracy may worsen using BN, due to inaccurate estimation of the statistics.

\subsection{Short-Term Adaptation}
\label{short-term-adaptation}
With short-term adaptation, we refer to an experiment that will only see the same test-time image once. In the next section, we will go into long-term adaptation, where we repeat the experiment multiple times in a row. In this section, we evaluate short-term adaptation by first evaluating the source model on the validation set of the source domains (in-distribution), then sequentially evaluating it on the target domains (out-of-distribution) and finally evaluating it on the source domains again (in-distribution). We evaluate on the validation set before and after adapting to the target domains, such that we can measure forgetting. Forget rate is the difference in accuracy before and after the continual adaptation to the target domains.

\subsubsection{Short-Term Adaptation to Contextual Shifts on CORe-50}
\partitle{Default train domains -> default test domains.}
In section~\ref{setup} we explain what the default train/test split is in CORe50. If we train a source model on these train domains, the model should generalize quite well on the test domains. Despite the minimal distribution shifts, we still evaluate the test-time adaptation techniques on this split, as they should work well regardless of the presence and strength of shift. Furthermore, these results can be compared to other literature using the same split.

The results for adapting from the default train domains to the default test domains from CORe50 are shown in Table \ref{core50-results-default}. Even though the test distribution was hypothesized to be relatively similar to the train distribution, BN still brings a significant improvement over the source model on sessions seven and ten. We see that TENT and CoTTA only bring small improvement over BN. We find that the small shift (and possibly short adaptation length) in this experiment will not lead to forgetting.

\begin{table*}[h]
    \caption{Classification accuracy~(\%) for continual test-time adaptation on CORe50 \cite{lomonaco2017core50} with source domains 1, 2, 4, 5, 6, 8, 9, 11.}
    \label{core50-results-default}
    \centering
    \scalebox{0.8}{
        \tabcolsep3pt
        \begin{tabular}{l|ccccc|c|c}
        \multicolumn{1}{l}{}& \multicolumn{5}{l}{$t\xrightarrow{\hspace*{4.2cm}}$}\\ \hline
        Method & \rotatebox[origin=c]{70}{1,2,4,5,6,8,9,11} & 3 & 7 & 10 & \rotatebox[origin=c]{70}{1,2,4,5,6,8,9,11} & Mean & Forget rate\\ 
        \hline
        Source & \textbf{100.0} & 83.6 & 54.0 & 59.0 & \textbf{100.0} & 79.3 & 0.0 \\
        BN \cite{schneider2020improving} & \textbf{100.0} & 83.6 & 68.4 & 66.9 & \textbf{100.0} & 83.8 & 0.0 \\
        TENT \cite{wang2020tent} & \textbf{100.0} & \textbf{84.7} & \textbf{69.5} & 66.9 & \textbf{100.0} & \textbf{84.2} & 0.0 \\
        CoTTA \cite{wang2022continual} & \textbf{100.0} & 83.7 & \textbf{69.5} & \textbf{67.3} & \textbf{100.0} & 84.1 & 0.0 \\
        \end{tabular}
    }
\end{table*}

\partitle{Outdoor -> indoor.}
The results for continually adapting from outdoor to indoor domains from CORe50 are shown in Table \ref{core50-results-outdoor}. We find that for each target domain, BN leads to an improvement of accuracy. Furthermore, TENT leads to an even more significant improvement on all target domains. However, TENT also results in forgetting, causing a small accuracy drop on the source domain after the adaptation sequence. Interestingly, although CoTTA does consistently outperform BN, it never outperforms TENT for any of the domains. We hypothesize CoTTA needs more iterations to improve on TENT, but it is impossible to see that from this experiment.

\begin{table*}[h]
    \caption{Classification accuracy~(\%) for short-term adaptation on CORe50 \cite{lomonaco2017core50} with domains 4, 10, 11 as source and effective batch size of 400.}
    \label{core50-results-outdoor}
    \centering
    \scalebox{0.8}{
        \tabcolsep3pt
        \begin{tabular}{l|cccccccccc|c|c}
        \multicolumn{1}{l}{}& \multicolumn{10}{l}{$t\xrightarrow{\hspace*{7.9cm}}$}\\ \hline
        Method & \rotatebox[origin=c]{70}{4,10,11} & 1 & 2 & 3 & 5 & 6 & 7 & 8 & 9 & \rotatebox[origin=c]{70}{4,10,11} & Mean & Forget rate\\ 
        \hline
        Source & \textbf{100.0} & 59.1 & 51.8 & 57.9 & 51.5 & 26.8 & 31.3 & 56.8 & 49.1 & \textbf{100.0} & 58.4 & 0.0 \\
        BN & \textbf{100.0} & 64.1 & 60.1 & 59.8 & 62.9 & 59.1 & 43.1 & 64.3 & 53.7 & \textbf{100.0} & 66.7 & 0.0 \\
        TENT \cite{wang2020tent} (lr=8e-4) & \textbf{100.0} & \textbf{68.8} & \textbf{68.0} & \textbf{71.5} & \textbf{73.3} & \textbf{67.5} & \textbf{55.8} & \textbf{73.9} & \textbf{71.3} & 98.1 & \textbf{74.8} & 1.9 \\
        CoTTA \cite{wang2022continual} (\textalpha=.99, p=.01) & \textbf{100.0} & 64.9 & 62.4 & 63.1 & 68.2 & 62.8 & 49.2 & 69.3 & 60.1 & 99.9 & 70.1 & 0.1 \\
        \end{tabular}
    }
\end{table*}

\partitle{Single domain -> all other domains.}
The results for continually adapting from a single domain to all other domains from CORe50 are shown in Table \ref{core50-results-single}. The findings are pretty similar to adapting from outdoor to indoor, however CoTTA does outperform TENT on some domains in this experiment. Another difference is that we see less forgetting in TENT and more in CoTTA, but nothing very significant.

\begin{table*}[h]
    \caption{Classification accuracy~(\%) for short-term adaptation on CORe50 \cite{lomonaco2017core50} with domain 1 as source (effective batch size is 400).}
    \label{core50-results-single}
    \centering
    \scalebox{0.8}{
        \tabcolsep3pt
        \begin{tabular}{l|cccccccccccc|c|c}
        \multicolumn{1}{l}{}& \multicolumn{12}{l}{$t\xrightarrow{\hspace*{9.6cm}}$} \\ \hline
        Method & 1 & 2 & 3 & 4 & 5 & 6 & 7 & 8 & 9 & 10 & 11 & 1 & Mean & Forget rate\\ 
        \hline
        Source & \textbf{100.0} & 47.0 & 51.8 & 20.9 & 49.7 & 30.7 & 35.3 & 64.6 & 51.0 & 13.8 & 23.9 & \textbf{100.0} & 49.1 & 0.0 \\
        BN \cite{schneider2020improving} & \textbf{100.0} & 62.8 & 64.3 & 53.6 & 60.2 & 51.9 & 48.4 & 74.7 & 61.9 & 45.0 & 52.0 & \textbf{100.0} & 64.6 & 0.0 \\
        TENT \cite{wang2020tent} (lr=3e-4) & \textbf{100.0} & \textbf{65.0} & \textbf{67.4} & 56.4 & 63.1 & \textbf{58.9} & \textbf{55.7} & \textbf{76.1} & 65.8 & \textbf{52.9} & 58.5 & 99.6 & \textbf{68.3} & 0.4 \\
        CoTTA \cite{wang2022continual} (\textalpha=.99, p=.01) & \textbf{100.0} & 64.5 & 66.3 & \textbf{57.8} & \textbf{63.2} & 57.5 & 54.0 & 74.7 & \textbf{66.1} & 52.0 & \textbf{59.2} & 99.2 & 67.9 & 0.8 \\
        \end{tabular}
    }
\end{table*}

\subsubsection{Short-Term Adaptation to Semantic Shifts on DomainNet}
\partitle{Real images -> paintings, cliparts and sketches.}
The results for continually adapting from real images to paintings, cliparts and sketches from DomainNet-126 are shown in Table \ref{domainnet-results-real}. We find that overall, BN performs comparable with source (with a very slight improvement overall), however for clipart specifically we see a small decrease in performance. TENT and CoTTA show only a small improvement over source. Noteworthy here is that CoTTA is showing double the forgetting. Updating all parameters in the model for CoTTA increases the effect of forgetting as opposed to only updating the batch normalization transformation parameters for TENT.

\begin{table*}[h]
    \caption{Classification accuracy~(\%) for short-term adaptation on DomainNet \cite{peng2019moment} with real images as source (effective batch size is 400).}
    \label{domainnet-results-real}
    \centering
    \scalebox{0.8}{
        \tabcolsep3pt
        \begin{tabular}{l|ccccc|c|c}
        \multicolumn{1}{l}{}& \multicolumn{5}{l}{$t\xrightarrow{\hspace*{3.5cm}}$}\\ \hline
        Method & \rotatebox[origin=c]{70}{Real} & \rotatebox[origin=c]{70}{Painting} & \rotatebox[origin=c]{70}{Clipart} & \rotatebox[origin=c]{70}{Sketch} & \rotatebox[origin=c]{70}{Real} & Mean & Forget rate\\ 
        \hline
        Source & \textbf{98.2} & 62.7 & 55.5 & 46.4 & \textbf{98.2} & 72.2 & 0.0 \\
        BN \cite{schneider2020improving} & \textbf{98.2} & 63.4 & 54.3 & 47.7 & \textbf{98.2} & 72.4 & 0.0 \\
        TENT \cite{wang2020tent} (lr=1e-4) & \textbf{98.2} & \textbf{65.1} & 57.0 & 53.2 & 95.9 & \textbf{73.9} & 2.4 \\
        CoTTA \cite{wang2022continual} (\textalpha=.99, p=.1) & 97.8 & 65.0 & \textbf{57.6} & \textbf{53.3} & 93.6 & 73.5 &  4.2 \\
        \end{tabular}
    }
\end{table*}

\partitle{Paintings -> real images, cliparts and sketches.}
The results for continually adapting from paintings to real images, cliparts and sketches from DomainNet-126 are shown in Table \ref{domainnet-results-painting}. BN leads to a bigger improvement over source than for adapting from real images, albeit still quite small. TENT does only very slightly outperform BN, but shows some forgetting. CoTTA outperforms TENT on all target domains, however, it shows even more forgetting. We hypothesize that the shift from paintings to the other domains is bigger than when shifting from real images to the other domains, as real images show much more detail and will probably have more variety. Paintings can be seen as reconstructions of the real world, therefore almost never being as rich in information as a real image. Another possible reason is that the target domains together account for 54\% more images when source is paintings instead of real images, making the adaptation longer, which could very well cause more forgetting as well. 

\begin{table*}[h]
    \caption{Classification accuracy~(\%) for short-term adaptation on DomainNet \cite{peng2019moment} with paintings as source (effective batch size is 400).}
    \label{domainnet-results-painting}
    \centering
    \scalebox{0.8}{
        \tabcolsep3pt
        \begin{tabular}{l|ccccc|c|c}
        \multicolumn{1}{l}{}& \multicolumn{5}{l}{$t\xrightarrow{\hspace*{3.5cm}}$}\\ \hline
        Method & \rotatebox[origin=c]{70}{Painting} & \rotatebox[origin=c]{70}{Real} & \rotatebox[origin=c]{70}{Clipart} & \rotatebox[origin=c]{70}{Sketch} & \rotatebox[origin=c]{70}{Painting} & Mean & Forget rate\\ 
        \hline
        Source & \textbf{97.5} & 75.0 & 53.0 & 47.3 & \textbf{97.5} & 74.1 & 0.0 \\
        BN \cite{schneider2020improving} & 97.4 & 75.2 & 54.4 & 52.7 & 97.4 & 75.4 & 0.0 \\
        TENT \cite{wang2020tent} (lr=1e-4) & 97.4 & 76.1 & 56.9 & 56.8 & 92.0 & \textbf{75.8} & 5.4 \\
        CoTTA \cite{wang2022continual} (\textalpha=.99, p=.1) & 97.3 & \textbf{77.8} & \textbf{59.4} & \textbf{57.8} & 84.0 & 75.3 & 13.4 \\
        \end{tabular}
    }
\end{table*}

\subsection{Long-Term Adaptation}\label{long-term-adaptation}

With long-term adaptation, we refer to an experiment that will see the same test-time image multiple times. In the previous section, we covered short-term adaptation, where we fix the length of the experiment by the number of unique test images. In this section, once again we start by first evaluating the source model on the validation set of the source domains (in-distribution), then sequentially evaluating it on the target domains (out-of-distribution) and finally evaluating it on the source domains again (in-distribution). Different from short-term adaptation however, to measure long-term adaptation, we repeat this cycle multiple times and refer to one cycle as an epoch. Even though we measure accuracy on the source data after every epoch, we first make a copy of the model under adaptation and use that to evaluate on the source data. This copy is thrown away after every epoch, so the actual model under adaptation never sees the source data again during the experiment. Also, much like in standard supervised training, we randomize the order of images in each target domain after every epoch, leading to different batches every epoch. We do not change the order of the target domains though, which is the same order as in the short-term adaptation experiments.

\begin{figure}[h]
  \centering
  \includegraphics[width=.45\columnwidth]{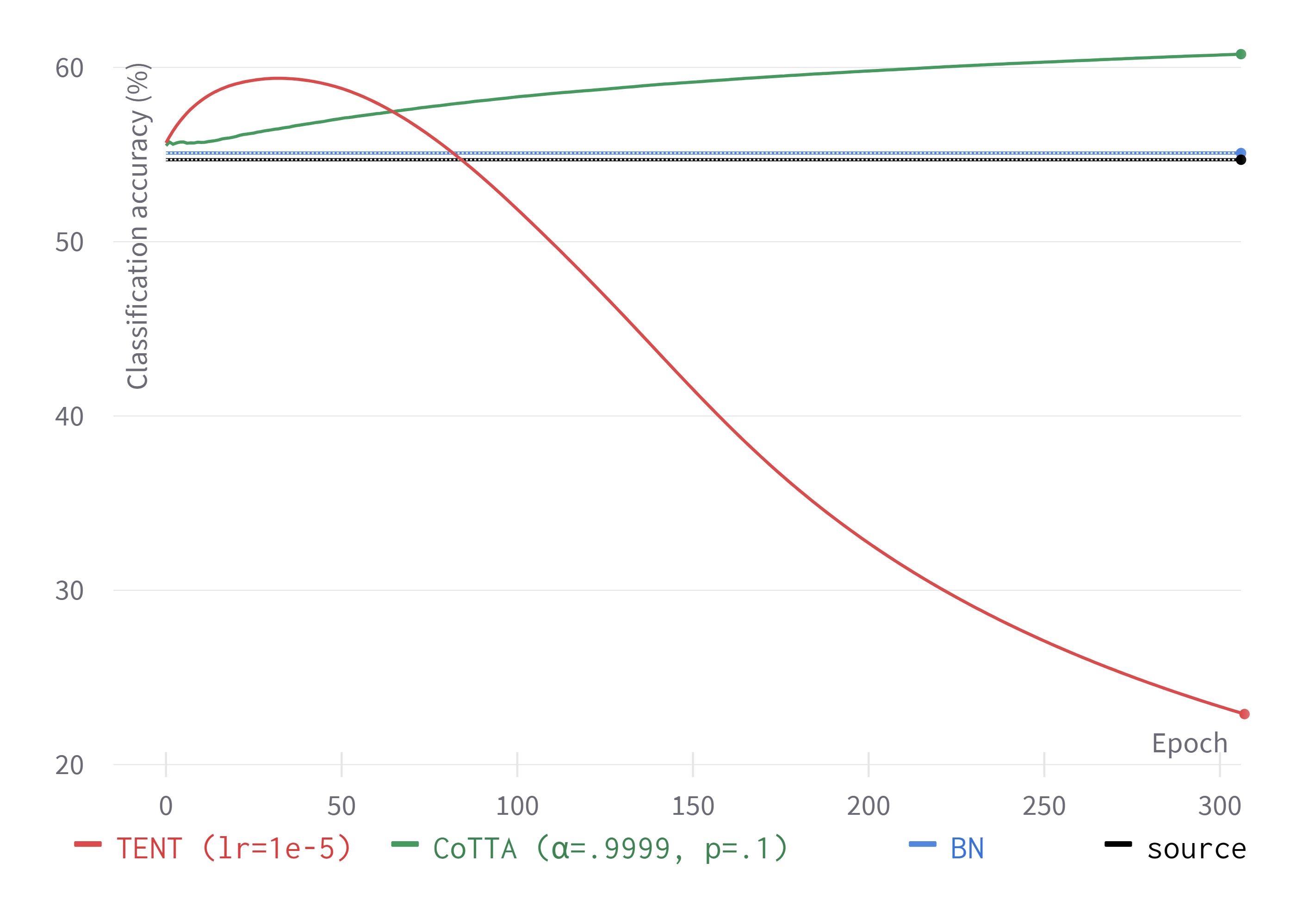}
  \includegraphics[width=.45\columnwidth]{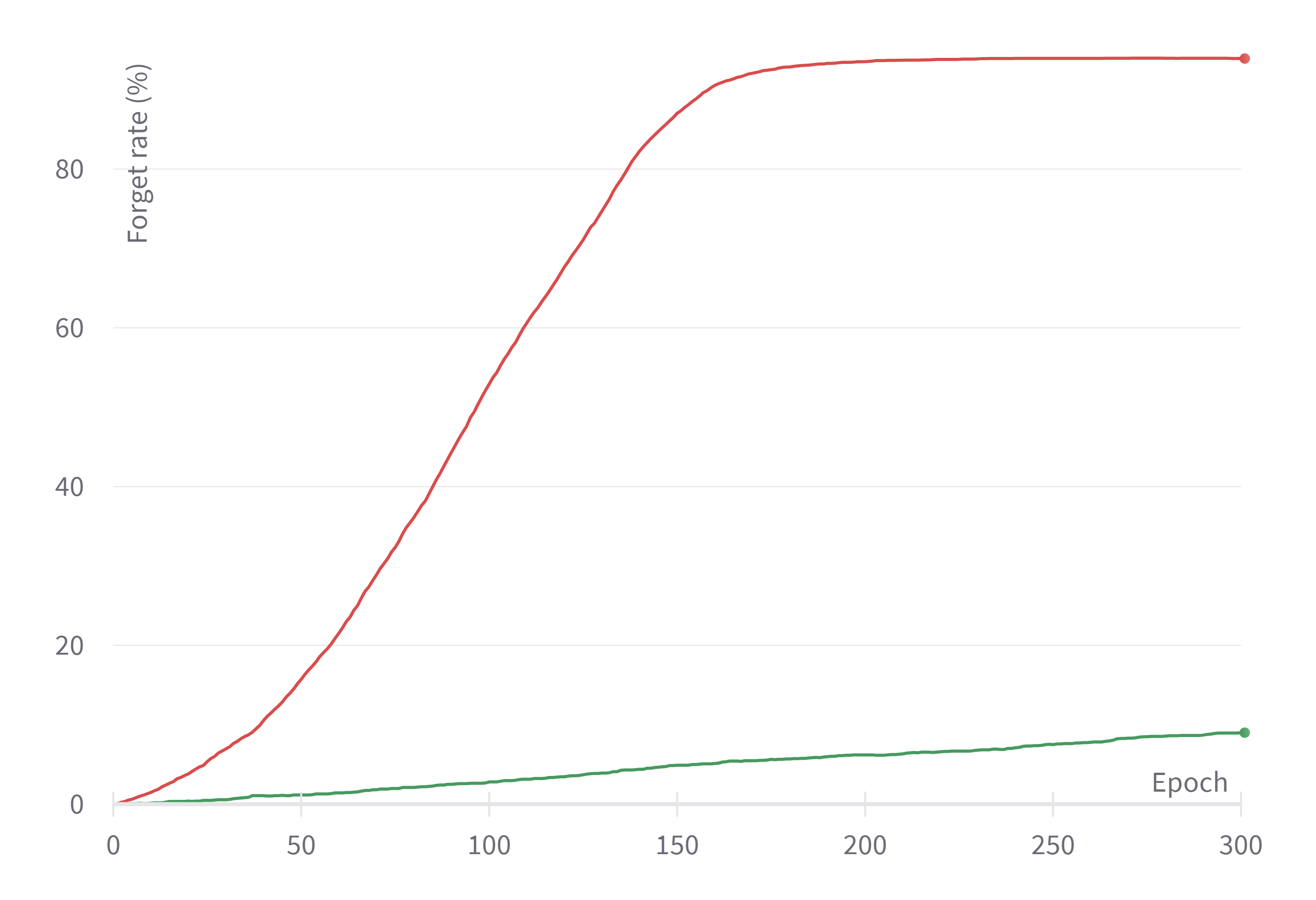}
  \caption{Long-term adaptation on DomainNet \cite{peng2019moment} with real images as source (effective batch size is 600).}
  \label{adapt-long-real-acc}
\end{figure}

The results for long-term adaptation from real images to paintings, cliparts and sketches from DomainNet-126 are shown in Figure \ref{adapt-long-real-acc}. We find that TENT quickly deteriorates in performance, whereas CoTTA keeps on improving. Also, TENT starts to forget more than linearly to the number of epochs, whereas for CoTTA forgetting is increased linearly and with relatively little amounts. We also see that CoTTA starts outperforming the maximum accuracy TENT can reach. Similar results have been found for the other source/target splits from section~\ref{short-term-adaptation}.

%% file: 5-discussion.tex
\section{Discussion}
This section will go into some very important points of discussion related to our work.

Once a model is deployed, it may be doing inference for an infinite amount of time. Unless we know beforehand on how much images a model will run inference on, we cannot derive much meaningful conclusions from experiments with a fixed length.

TENT is very sensitive to its learning rate. We had to carefully tune it to the specific experiment and its length to work well. CoTTA works well on the default learning rate, however also for CoTTA the other hyperparameters had to be tuned to the specific experiment and its length to work well. This may not be possible in a real life scenario, where you do not know what type of shifts will happen or how long your model will be running inference for.

The short-term-adaptation experiments can be quite deceiving, leading one to believe that TENT is better than CoTTA. However, it seems CoTTA is better at long-term adaptation and just needs more iterations to start working. Therefore, in some of the short-term experiments, we set the alpha hyperparameter quite aggressively, causing CoTTA to show a lot of forgetting, whereas it would not have such extreme forgetting in a long-term setting with a lower alpha.

From the limited long-term-adaptation experiments, we cannot conclude whether CoTTA will maintain its high accuracy forever. This requires further work, as described in~\ref{future-work}. It could be that by increasing the alpha hyperparameter, we are just postponing the error accumulation causing the performance to decrease. That would mean we are still tuning it to the specific length of the experiment and there is no way to flatten the performance curve somehow.

Also, we did not try lower learning rates of TENT for the long-term adaptation experiments. In order to be sure that TENT cannot perform as well as CoTTA in the long-term, we therefore suggest in \ref{future-work} to do more experiments on it.

In our experiments, we do not reset TENT (or CoTTA) whenever the target domain changes. In \cite{wang2022continual}, the authors show that this will improve performance significantly. However, to stay closer to a real life scenario where information about domain changes in unavailable, we opt not to reset the model.

We turned off augmentations in CoTTA because we find that the augmentations provided by the authors significantly worsen performance on our experiments. We hypothesize this could be because the source models weren't trained with such augmentations either. Experiments we did on other datasets show that using the same augmentations used during training for CoTTA does lead to performance improvement.

%% file: 6-conclusion.tex
\section{Conclusion}

In general, we find that the test-time adaptation methods show relatively more improvements on contextual shifts compared to semantic shifts. Also we find that the test-time-adaptation methods show more forgetting in semantic shifts compared to contextual shifts.

For relatively short fixed experiments lengths and carefully tuned hyperparameters for the specific experiment, we find that TENT overall outperforms the source model, BN and CoTTA.

From the (limited) long-term adaptation experiments, we find that CoTTA starts to outperform TENT when the experiment length is longer. These results indicate that CoTTA is better suited to a real-life scenario where a model is running inference for an infinite amount of time. We also find that the bigger the shift or the longer the experiment, the more forgetting occurs for both TENT and CoTTA. Forgetting seems to be linearly related to experiment length. 

Finally, we can conclude that, BN, given a large enough test batch size, will perform on par with or outperform the source model (with only one small exception in our experiments). On top of that, BN does not show error accumulation or catastrophic forgetting, as it is not optimization based and will produce the same results on the same test batch irrespective of the previous test batches. The gains of BN seem to be lower when the source model was trained on a dataset with a relatively wide distribution. 

%% file: 7-futurework.tex
\section{Future Work}\label{future-work}
There are quite some areas of further work that are very important to pursue, in order to derive more sound and insightful conclusions. This section will go into some areas of future work in order of importance. 
\begin{enumerate}
    \item For the long-term adaptation experiments, different versions of TENT should be added to the graph with different learning rates, in order to conclude whether TENT cannot perform as well as CoTTA in the long-term.
    \item The long-term adaptation experiments should be extended to even more epochs to find out whether CoTTA flattens or starts decreasing after a while. 
    \item Long-term adaptation versions should be added for the other experiments in the short-term experiments section.
    \item Add experiments with other splits of source and target domains, such that we can base our conclusions on more results.
    \item Do the batch size analysis also on DomainNet, as we saw examples where BN decreases performance which can almost not be explained by the very high effective batch size of 400. If there is another reason, find out what that is.
    \item Do the qualitative analysis also on DomainNet.
    \item Add experiments where the domain is shifting gradually instead of very abrupt distinct changes.
    \item Turn on augmentations for CoTTA and analyze when which augmentations work and how this is related to the augmentations used for source model training. Augmentations may further improve CoTTA's long-term adaptation performance, as every epoch CoTTA will see different augmentations of the same image, much like normal supervised training.
\end{enumerate}